\documentclass{article}
\usepackage{cite}
\usepackage{booktabs}
\usepackage{amsmath,amssymb,amsfonts}
\usepackage{algorithmic}
\usepackage{float}
\usepackage{graphicx}
\usepackage{spconf}
\usepackage{subcaption}
\usepackage{textcomp}
\usepackage{todonotes}
\usepackage{xcolor}
\usepackage{xurl}

\title{Investigating Modality Contribution in Audio LLMs for Music}
\name{Giovana Morais$^\dagger$ \qquad Magdalena Fuentes$^{\dagger,\star}$}
\address{$^\dagger$ Music and Audio Research Lab, New York University, USA\\
$^\star$ Integrated Design \& Media, New York University, USA}

\begin{document}
\newcommand{\gi}[1]{\textcolor{teal}{#1}}
\newcommand{\change}[1]{\textcolor{red}{#1}}

\ninept
\maketitle
\begin{abstract}
Audio Large Language Models (Audio LLMs) enable human-like conversation about
music, yet it is unclear if they are truly listening to the audio or just using
textual reasoning, as recent benchmarks suggest. This paper investigates this
issue by quantifying the contribution of each modality to a model's output.
We adapt the MM-SHAP framework, a performance-agnostic score
based on Shapley values that quantifies the relative contribution of each
modality to a model’s prediction. We evaluate two models on the MuChoMusic
benchmark and find that the model with higher accuracy relies more on text to
answer questions, but further inspection shows that even if the
overall audio contribution is low, models can successfully localize key sound
events, suggesting that audio is not entirely ignored. Our study is the first application of MM-SHAP to Audio LLMs and we hope it will serve as a foundational step for future research in explainable AI and audio.
\end{abstract}


\section{Introduction}\label{sec:introduction}

Audio Large Language Models (Audio LLMs) aim to expand the capabilities of LLMs
by incorporating audio information into their
reasoning~\cite{deshmukh2024pengiaudiolanguagemodel}. While the number of
proposed models continues to grow
\cite{chu2023qwenaudio,liu2023mullama,deng2024musilingo,gong2024ltu,tang2024salmonn}
difficulty lies in assessing free-form text outputs, due to the unstructured
nature of the predictions~\cite{liang2023holisticevaluationlanguagemodels}.
Additionally, evaluation benchmarks must go beyond measuring natural language
understanding to explicitly test a model’s ability to reason using audio.
Multiple evaluation datasets have been proposed that assess model performance in
tasks such as captioning \cite{agostinelli2023musiccaps} and compositional
reasoning \cite{ghosh2024compa}. The lack of consistent evaluation methods
inspired researchers to build new comprehensive benchmarks to evaluate and
compare Audio LLMs, such as MuChoMusic \cite{weck2024muchomusic} and the Massive
Multi-Task Audio Understanding and Reasoning Benchmark (MMAU)
\cite{sakshi2024mmau}.  These datasets feature multiple-choice questions, with
the goal that the model will choose the most appropriate answer based on the
input audio. The main advantages of this evaluation method are scalability and
simplicity, since the model's answers are matched against predefined options
rather than parsed from open-ended text.  Furthermore, the questions are
designed to test various dimensions of music understanding, such as historical
knowledge, music structure, and rhythm.

A surprising outcome is that audio-text models perform poorly on these benchmarks, despite achieving strong results on tasks such as audio captioning and description \cite{weck2024muchomusic,sakshi2024mmau}, although the reasons for this remain unclear.
Experiments that replace the input audio with random noise or silence result in negligible changes to model performance \cite{weck2024muchomusic}, suggesting that the models' use of the audio modality may be limited.
While these outcomes indicate a potential imbalance in the modalities (i.e., the text modality is the dominant contributor to the model's output), the benchmarks primarily rely on indirect metrics, like accuracy, which can be misleading.
Accuracy-based evaluations fail to capture cases where a model uses relevant information from a modality but still produces an incorrect prediction.

Similar modality-imbalance issues have been reported in Vision LLMs, where unimodal models can often achieve
comparable accuracy to a multimodal one on a vision-language task~\cite{gat2021perceptual},
suggesting unimodal collapse. To better understand this phenomenon without relying on performance metrics, \cite{parcalabescu2023mmshap,parcalabescu2024measuring}
proposed MultiModal-SHAP (MM-SHAP). They utilize Shapley values \cite{shapley1953}, a post-hoc explainability technique to
quantify the relative contribution of each modality in a model’s prediction,
irrespective of its accuracy.
We investigate MM-SHAP's adaptability to the modality of audio. Particularly, our main contributions are that we:
\begin{enumerate}
    \item Adapt MM-SHAP \cite{parcalabescu2023mmshap} to inspect how Audio LLMs are using each modality in different tasks;
    \item Investigate two well-known Audio LLMs (Qwen-Audio~\cite{chu2023qwenaudio} and MU-LLaMA \cite{liu2023mullama}) in multiple-choice Q\&A using the open-source MuChoMusic benchmark dataset \cite{weck2024muchomusic}; and
    \item Examine how the Audio LLMs use the two modalities. We show that the usage of text is higher for multiple-choice questions, aligning with results from Vision LLMs. We also demonstrate that good performance on MuChoMusic does not imply balanced modality contributions and vice versa.
\end{enumerate}

Our aim is to gain insight into how much these models employ each modality to reason over music, and how the task and the MM-SHAP design choices affect these interactions.


\section{Related Work}
\label{sec:related_work}

\begin{figure*}[t]
    \centering
    \includegraphics[width=\linewidth]{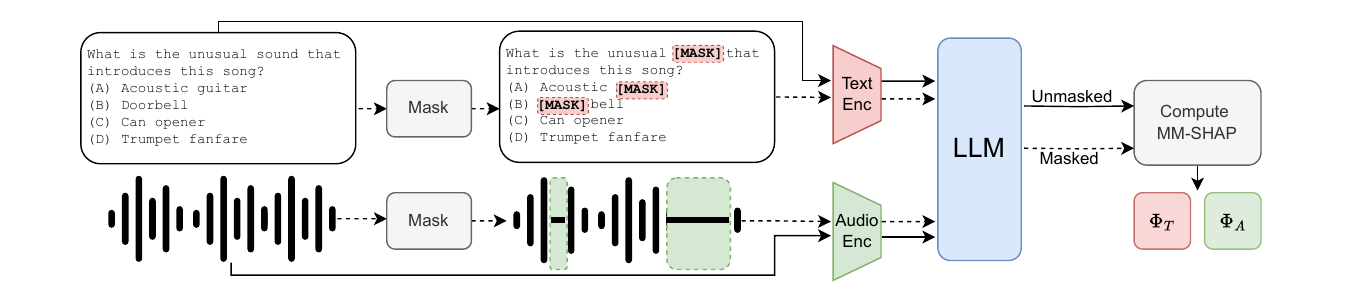}
    \caption{We compute Shapley values by masking all combinations of inputs (approximated via random permutation), and averaging the change in the logits of the base answer (unmasked inference indicated in solid line). We mask text tokens and audio waveform segments.}
    \label{fig:experiment_setup}
\end{figure*}

Within explainability techniques in machine learning, there is a family of
``post-hoc" methods whose aim is to analyze how each input feature contributes
to a given model output via input perturbations. These include approaches like
LIME (Local Interpretable Model-agnostic Explanations)~\cite{ribeiro2016lime},
that explain a prediction by approximating it locally with an interpretable
model (e.g., linear regression or a decision tree). To the best of our
knowledge, most of the post-hoc techniques applied to music use LIME or
LIME-variations \cite{haunschmid2020audiolime},\cite{mishra2017slime}.
Recently, \cite{sotirou2024musiclime} expanded LIME methods to investigate music
understanding models by analyzing the most important features for genre
classification from both text and audio modalities. As for modality contribution,
the DIsentangled Multimodal Explanations (DIME)~\cite{lyu2022dime} was
proposed as a visualization technique to gain insights into each modality's
contribution to the model's output. DIME disentangles a model into unimodal and
multimodal contributions and then produces visualizations on each sub-model
using LIME.  Because of its focus on visual explanations, DIME is hard to use as
a quantitative measure of modality contribution.

A common strategy for assessing modality contribution is to conduct
ablation studies by removing or replacing one of the modalities with a random
input. For example, \cite{weck2024muchomusic} swaps the audio with silence,
white noise, or another audio and evaluates how much the accuracy changes.
The main limitation of this approach is that completely removing a modality
obscures how the model uses each modality and their interactions, and to what
extent its output depends on the missing modality. As argued in
\cite{hessel2020doesmymodel}, comparing the unimodal and multimodal
performances does not explain how the model uses cross-modality interactions.

To address the limitations of ablation studies and provide stronger theoretical
guarantees than surrogate models like LIME, another family of ``post-hoc''
methods based on Shapley values has gained prominence. Shapley values are the
only feature attribution method that satisfies theoretical guarantees
\cite{molnar2025interpretable}.
To the best of our knowledge, the only exploration of Shapley values and music
audio was performed in \cite{bolanos2025benchmarkingexplanations}.
Recently, Parcalabescu and Frank proposed MM-SHAP~\cite{parcalabescu2023mmshap}, which quantifies modality contribution
in text-vision models by adapting Shapley values to measure how input
perturbations across modalities affect the model’s output.


\section{Method}
\label{sec:method}

\subsection{Shapley Values and Feature Contribution}
\label{ssec:shapley_values}
Shapley values were first proposed in the context of game theory to estimate how
much each player contributes to the overall outcome of a cooperative
game~\cite{shapley1953}. They were adapted for explaining machine learning
methods by~\cite{lundberg2017shap} in the SHAP (SHapley Additive exPlanations) framework, where features act as players contributing
to a model's output. The Shapley value $\phi$ for a feature $j$ and token $t$ is
given by

\begin{equation}\label{eq:shap_single}
    \phi_{j,t} = \sum_{S \subseteq \{1, \cdots, n\} /\{j\}} \frac{f_t(S\cup\{j\}) - f_t(S)}{\gamma},
\end{equation}

\noindent where $S$ represents subsets of features excluding the feature $j$, $f_t(S)$ is
the model's output for the subset $S$ and token $t$, $n$ defines the total
number of features.  Therefore, $\phi_{j,t}$ represents how much feature $j$
contributes to the output token $t$. 
The normalizing factor $\gamma = \frac{n!}{|S|! (n-|S|)!}$ weights the marginal
contribution of each feature fairly across all possible subsets of features. A
subset's marginal contribution is weighted proportionally to how often that
subset appears when averaging over all permutations. Shapley values satisfy the
theorical guarantees of symmetry, efficiency, dummy, and additivity, which
together constitute a fair payout
\cite{molnar2025interpretable}.

The exact computation of Shapley values requires evaluating all possible
subsets of features. However, as the number of features grows, the computation of
Equation~\ref{eq:shap_single} has an exponential time complexity of $O(2^n)$.
This means that, for machine learning models with a large numbers of
features, the exact computation of Shapley values is too expensive.
In practice, Shapley values are approximated by sampling feature subsets, e.g., using KernelSHAP or PermutationSHAP from the SHAP framework~\cite{lundberg2017shap}.
In this work, we use PermutationSHAP, which approximates Shapley values by
averaging each feature’s marginal contribution across random permutations,
measuring the change in the model's output when the feature is added to
preceding subsets. When estimated using $m$ randomly sampled subsets,
PermutationSHAP has a complexity of $O(m*n)$. We use the default value of
$m=10$ from the \texttt{shap} library.

\subsection{Features and Masking}
For audio we mask the raw waveform by zeroing out short waveform segments. The advantage of masking raw inputs rather than intermediate audio representations is that it prevents information leakage (e.g., in case the audio tokens are computed with temporal overlap or redundancy), and allows for further interpretability of how the masked parts of the input affect the output.

Our audio masking adapts dynamically to text length, aiming for a balanced
number of features from each modality. This is accomplished by adjusting the
window size such that the quotient of dividing the audio length by the number of
windows equals the number of text tokens ($n_A = n_T$), following
\cite{parcalabescu2023mmshap}.  For example, in a 10-second audio clip and
100 text tokens, the masked windows are approximately 100ms long.

When masking text tokens, we replace them with the [MASK] token. We do not mask certain special tokens such as ``audio indicators'' which signal that audio features follow (\verb|<audio>|, \verb|<audio_padding>|, \verb|#Audio|), and question and answer indicators (\texttt{\textless|question|\textgreater}, \texttt{\textless|answer|\textgreater}).

\subsection{Computing Model Scores}
For a classification task such as the ones explored in~\cite{parcalabescu2023mmshap}, the model's score for feature subset $S$ given a class $c$ is defined as the output probability of the model $f_c(S)$ for that class. The Shapley values are then computed by measuring how the probabilities change as the input features are masked. Scores can be computed at the class-level or averaged across classes for a ``global'' effect.
For a generation task, this approach would be too costly as the ``classes'' involved span the whole token vocabulary of the LLM. Instead, we leverage the ideas in~\cite{parcalabescu2025do} to develop the methodology illustrated in Figure \ref{fig:experiment_setup}. First, we obtain a set of tokens $T$ corresponding to the model's answer \textit{without} masking any modality (i.e., ``(B) Doorbell'' in Figure \ref{fig:experiment_setup}). We save the indices of the
tokens of the generated answer, and then compute $f_t(S)$ by summing the changes in the logits for each token $t \in T$ after masking parts of the input (in both text and audio). This way we measure the changes of the model with respect to the baseline, unmasked answer. 

\begin{figure*}[t]
    \centering
    \includegraphics[width=\linewidth]{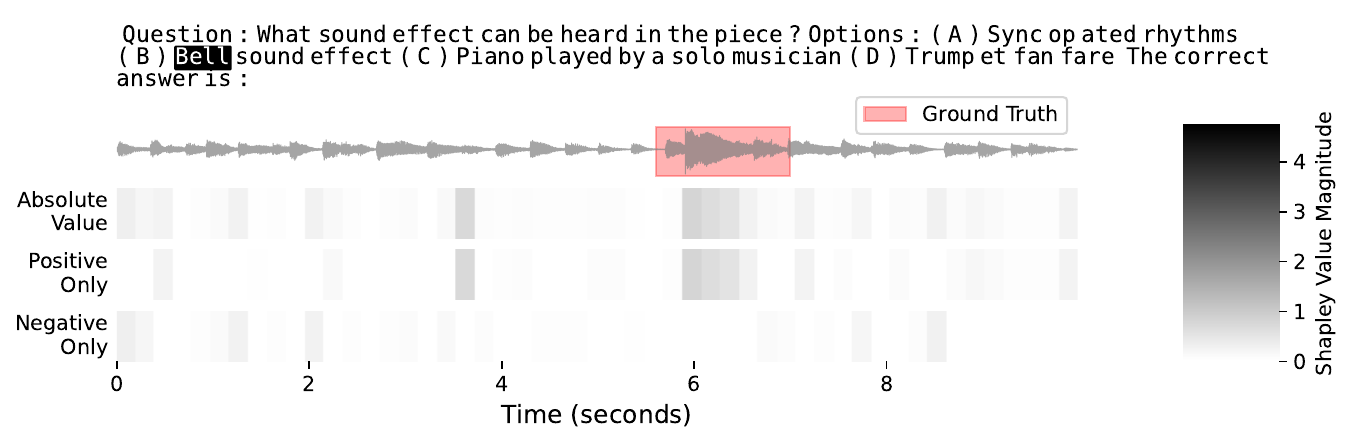}
	\caption{QwenAudio MC-NPI results for MusicCaps track
		\texttt{QK-mjNg8cPo}: the output is ``The sound effect that can be
		heard in the piece is a bell sound effect". The image shows the modality contribution for output token ``bell", i.e., $\Phi_{A,t}$ and $\Phi_{T,t}$, where $t = \text{bell}$.
        The top section highlights the most important text tokens from the input question (black = highest contribution, with a threshold of 80\% of the maximum Shapley value for readability). The bottom section shows the audio waveform with its corresponding Shapley value contributions: absolute value, positive, and negative components. Darker colors mean higher contribution to the model's output token. }
    \label{fig:comparison_qwenaudio_zs}
\end{figure*}

\subsection{Modality Contribution}
While Equation \ref{eq:shap_single} captures the impact of a single feature $j$, we are interested in understanding how sets of features, grouped by modality, contribute to the model's output. Following the proposal of~\cite{parcalabescu2023mmshap}, we define each modality contribution as
\begin{align}
    \Phi_A = \sum_t^T\sum_{j=1}^{n_A} |\phi_{j,t}| &  &
    \Phi_T= \sum_t^T\sum_{j=n_A+1}^{n_A+n_T} |\phi_{j,t}|,
\end{align}

\noindent where $\Phi_A$ and $\Phi_T$ represent the scores of all audio and text features,
and $n_A$ and $n_T$ are the number of audio and text tokens, respectively. We
note here that audio and text features are concatenated.
Shapley values measure positive and negative contribution amounts. However, for measuring modality contribution, we are interested in \textit{how much} a feature contributed to the expected output. Therefore, we use the absolute value of the Shapley value $\phi_{j,t}$.
To directly compare modality influence, we define T-SHAP (Textual SHAP) and A-SHAP (Audio SHAP) as the normalized ratios of each modality’s total contribution. For a perfectly balanced model, both T-SHAP and A-SHAP would equal $0.5$.

\begin{align}
    \text{A-SHAP} = \frac{\Phi_A}{\Phi_T + \Phi_A } & & \text{T-SHAP} = 1 - \text{A-SHAP}.
\end{align}

\section{Experiments}
\label{sec:experiments}

\noindent\textbf{Data}
Our experiments focus on the MuChoMusic benchmark \cite{weck2024muchomusic}, as
we have access to both the audio and the questions' answers. MuChoMusic is made
of 1,187 multiple-choice questions validated by human annotators and associated
with 644 music tracks sourced from the SongDescriberDataset (SDD) and MusicCaps.
We limit our experiments to MusicCaps tracks to make a fair comparison between models, since Qwen-Audio focuses on the first 30s of the audio\footnote{\url{https://github.com/QwenLM/Qwen-Audio?tab=readme-ov-file#quickstart}}, while MU-LLaMA processes entire tracks. 
MusicCaps questions constitute
approximately 71\% of MuChoMusic.\\

\noindent\textbf{Models}
Our experiments compare two Audio LLMs, which represent the best and second-worst performing models on MuChoMusic. MU-LLaMA~\cite{liu2023mullama} uses MERT-v1-330M 
 as the audio encoder and
    LLaMA 2 7B 
	 as the language encoder, then trains an adapter
     to map the MERT output to the language model. All audio
    is resampled to 24 kHz before encoding. The model is trained with the MusicQA dataset, which is
    compiled from the MusicCaps~\cite{agostinelli2023musiccaps} and
MagnaTagATune~\cite{law2009magnatagatune} datasets.
	Qwen-Audio~\cite{chu2023qwenaudio} uses Whisper-large-v2 as its base audio
	encoder and Qwen-7B as its base LLM.
    Audio is resampled to 16 kHz prior to encoding.
	We use Qwen-Audio-Chat for our
	experiments as it is fine-tuned to answer questions. \\

\noindent\textbf{Experiments}
To explore the effect of the number of text tokens on the audio usage, we investigate the models in two different setups: \\
\begin{enumerate}
	\item{\textbf{Multiple-Choice with Previous Instructions (MC-PI):} We use the original input
       of MuChoMusic containing an in-context example
       followed by the actual multiple-choice question we want the model to
	   answer.}
	\item{\textbf{Multiple-Choice with No Previous Instructions (MC-NPI):} We
		omit the in-context example (i.e., fewer text tokens).}
\end{enumerate}

For each task and model, we report A-SHAP scores
and accuracy, calculated with the code provided in the MuChoMusic GitHub
repository. 
We kept all system instructions with their default values (``You're a
reliable assistant, follow these instructions.").
Unlike~\cite{parcalabescu2023mmshap}, we did not mask system instructions, only the
question itself.

We also performed a qualitative analysis to investigate how the Shapley
values behave for questions requiring single sound event vs. those requiring longer-term audio understanding to generate the answer. We looked at each model and each
experiment, filtered only questions with correct answers and selected 5 samples with the highest A-SHAP values, and 5 with
the lowest. We removed duplicates, a sample with corrupted audio, and a question with no correct distractor, resulting in 35 unique Q\&A.
From this set, we first identified questions answered by a single sound event.
For these, we manually annotated the relevant audio segment, as in
Figure~\ref{fig:comparison_qwenaudio_zs}. We observed that most of these
questions still required a long-term understanding of the audio. To find more
single-sound-event examples, we searched for keywords that were common in such
questions (e.g., ``interrupted'', ''at the beginning``, ''sound at the end``). We
identified a total of 13 Q\&A pairs.
All of our experiments and
code are available on our GitHub\footnote{\url{https://github.com/giovana-morais/2025_investigating_mmshap}}.

\section{Results and Discussion}
\label{sec:results}

\begin{table}[h]
\centering
\vspace{-6pt}
\caption{Accuracy and average A-SHAP for the two experiments.}
\label{tab:a_shap}
\begin{tabular}{@{}ccccc@{}}
\toprule
\textbf{} & \multicolumn{2}{c}{\textbf{Accuracy}} & \multicolumn{2}{c}{\textbf{A-SHAP}} \\ \midrule
 & \textbf{MC-PI} & \textbf{MC-NPI} & \textbf{MC-PI} & \textbf{MC-NPI} \\ \midrule
MU-LLaMA & $0.30$ & $0.32$ & $0.50 \pm 0.02$ & $0.47 \pm 0.02$ \\
QwenAudio & $0.44$ & $0.47$ & $0.23 \pm 0.02$ & $0.21 \pm 0.02$ \\ \bottomrule
\end{tabular}
\end{table}

\noindent Table~\ref{tab:a_shap} shows that the more accurate model, Qwen-Audio, relies
less on audio, whereas MU-LLaMA uses both modalities in a balanced manner. Indicating that there is no relationship between accuracy and modality usage.
This was further reinforced by analyzing A-SHAP values grouped by correct and incorrect questions, which showed no trend.
Examining questions broken down
by knowledge dimension according to MuChoMusic categories (e.g. knowledge, reasoning) also showed no significant effect in the A-SHAP value.
We think that this lack of difference is consistent with the findings in
\cite{zang2025reallylisteningboostingperceptual}, where the authors show that LLMs perform as
good on MuChoMusic as Audio LLMs, indicating that the audio modality is not
leveraged as expected in the answering process. They show that this is caused by
MuChoMusic approach to designing the multiple-choice distractors
which, as a result, any LLM capable of eliminating the two unrelated
answers can achieve an accuracy of 50\% by randomly selecting the answer from the
two remaining options. What the results of
\cite{zang2025reallylisteningboostingperceptual} suggest is that what
MuChoMusic is, in fact, testing
the reasoning skills of an LLM model instead of the perceptual capabilities of
Audio LLMs. To verify whether the low audio usage is extended to tasks other than multiple-choice, we additionally evaluated the best-performing model (Qwen-Audio) by asking it to describe the audio clips, with the same number of audio tokens as MC-PI, for fairness.
 The average A-SHAP increased to 0.73\%, suggesting stronger audio usage and significant task influence.

As we have seen little change in the audio
usage between correct and incorrect examples, we decided to further explore the
positive and negative contributions ($\phi_{j,t}$), together with the absolute
value that is in fact used to compute MM-SHAP. We analyze the songs as
described in Section~\ref{sec:experiments}. From the qualitative analysis, we
found that in Qwen-Audio the audio features have much lower magnitude than text features,
which aligns with the low audio usage. But this does not mean that the audio is
being discarded. For each specific token $t \in T$, we see that the positive contributions often align
with the audio region that contains the ground truth, and also with the
input token that represents that event, as shown in
Figure~\ref{fig:comparison_qwenaudio_zs} for the token ``bell''.
We note that the same behavior was found in
the other questions with single-sounding events.
It is also possible to see other strong
activations, and those are not related to the correct answer
(in
Figure~\ref{fig:comparison_qwenaudio_zs}, they represent a piano note). We hypothesize that in
such cases, the information is useful for the model's internal mechanisms, but not
necessarily are plausible to the data explanation. In examples that require
longer-term explanations, the activations are spread throughout the audio.
In MU-LLaMA, the audio and text features had similar magnitude overall, but we
noticed that the audio features with high activations were spread throughout the
audio in long-term and single-sounding questions. Although there were
positive contributions in the correct regions, they were not as concentrated as in
Qwen-Audio. Also, MU-LLaMA activations showed negative contributions as
strong as the positive, which affect the final A-SHAP value that takes into
account only the magnitude of the feature importance.
We provide additional interactive plots comparing all models and experiments in the project's demo
page.

To the best of our knowledge, this is the first time modality contribution is measured in Audio LLMs, as inspected through Shapley values.
In this work, we followed the original proposal by \cite{parcalabescu2023mmshap}
closely. However,
we discuss below the factors that, in our opinion, limit the application of MM-SHAP for Audio LLMs as is, and venues for improvement and future research.

A primary challenge is the lack of ground truth for qualitative analysis.
Our analysis was restricted to questions with easily localizable sound events;
annotating questions about melody, instrumentation, or structure is harder, yet
those question types are the most frequent when evaluating music understanding
systems, which makes reliable annotation an open question by itself.
Furthermore, the audio segment size, which is tied to the number of text tokens,
may be too small to contain perceptually meaningful information, complicating
our cross-experiment comparison. For the audio masking, we zeroed the audio segments, but
other alternatives include replacing them with white noise or an average between
neighbor segments, as in \cite{bolanos2025benchmarkingexplanations} and \cite{mishra2020slimexploration}.
Determining an optimal audio window size and the optimal audio mask are
open research problems, especially if accounting for temporal \textit{and}
spectral information. This is especially important because the chosen mask must
be a ``neutral'' representation of the data.
While MM-SHAP and Shapley values can help us understand how much audio is used in current models,
the tool is not enough to hint us about how the audio is used internally in the
model, as is the case with post-hoc explanation methods.

\section{Conclusion}

In this work, we adapt MM-SHAP to measure modality contributions in Audio LLMs, analyzing how audio and text are used in perception tasks.
Through a systematic evaluation of Qwen-Audio and MU-LLaMA, we discovered that the best-performing model, Qwen-Audio, relied significantly more on its text modality.
While text processing appears to drive overall performance, our qualitative analysis confirms that the audio modality is not discarded, suggesting a gap between \textit{listening} to the audio and \textit{reasoning} about it.
Our findings also suggest that balanced modality use does not guarantee superior performance, challenging the assumption that optimal multimodal models must use all modalities equally. Future work will systematically study, at scale, how audio information is leveraged, how it survives multimodal integration, and how task formulation (e.g., multiple choice vs. descriptive) influences its use.

\pagebreak

\section{Acknowledgments}
This work is supported by the National Science Foundation under Grant Numbers 2504642 and 2504643. This work was supported in part through the NYU IT High Performance Computing resources, services, and staff expertise.

\bibliographystyle{ieeetr}
\bibliography{refs}

\end{document}